\documentclass{article}

     \PassOptionsToPackage{numbers, compress}{natbib}

    \usepackage[final]{neurips2020_preregistration}

\usepackage[utf8]{inputenc} 
\usepackage[T1]{fontenc}    
\usepackage{hyperref}       
\usepackage{url}            
\usepackage{booktabs}       
\usepackage{amsfonts}       
\usepackage{nicefrac}       
\usepackage{microtype}      
\usepackage{graphicx}
\usepackage[ruled,linesnumbered]{algorithm2e}
\usepackage{algorithmic}
\usepackage{float}

\title{A Proposal to Study "Is High Quality Data All We Need?"}

\author{%
  Swaroop Mishra \\
  \texttt{srmishr1@asu.edu} \\
   \And
   Anjana Arunkumar \\
   \texttt{aarunku5@asu.edu} \\
\And\\School of Computing, Informatics, and Decision Systems Engineering\\
    Arizona State University\\

}

\begin{document}

\maketitle

\begin{abstract}




Even though deep neural models have achieved superhuman performance on many popular benchmarks, they have failed to generalize to OOD or adversarial datasets. Conventional approaches aimed at increasing robustness include developing increasingly large models and augmentation with large scale datasets. However, orthogonal to these trends, we hypothesize that a smaller, high quality dataset is what we need. Our hypothesis is based on the fact that deep neural networks are data driven models, and data is what leads/misleads models. In this work, we propose an empirical study that examines how to select a subset of and/or create high quality benchmark data, for a model to learn effectively. We seek to answer \textit{if big datasets are truly needed to learn a task}, and whether a \textit{smaller subset of high quality data} can replace big datasets. We plan to investigate both \textit{data pruning} and \textit{data creation} paradigms to generate high quality datasets.

\end{abstract}

\section{Introduction}
\vspace{-2mm}
Deep neural models such as EfficientNet-B7 \cite{tan2019efficientnet}, BERT \cite{devlin2018bert} and RoBERTA \cite{liu2019roberta} have achieved super-human performance on many popular benchmarks in various domains such as Imagenet \cite{russakovsky2015imagenet}, SNLI \cite{bowman2015large}, and SQUAD \cite{rajpurkar2016squad}. However, their performance drops drastically on exposure to out of distribution (OOD) and adversarial datasets \cite{jia2017adversarial, eykholt2018robust, jin2019bert, hendrycks2020pretrained}. Lots of resources and time are being invested in developing better models and architectures, such as transformer based approaches \cite{vaswani2017attention}, that dominate leaderboards. \textit{Since deep learning --a data driven approach-- finds representation from data, shouldn't the focus be placed on creating `better' datasets rather than developing increasingly complex models?}\\
Let us consider this through an analogy-- a student ($A$) is asked to self-learn a concept by going through a question bank ($Q_1$), where there are 1000 solved questions. After self-learning, $A$ is tested using 100 unsolved questions present at the end of the $Q_1$. While $A$ achieves unprecedented performance (85/100), beating other students who are explicitly taught the concept, when tested on another 100 questions on the same topic from question bank ($Q_2$), $A$ fails on 50 questions. Similarly, if $A$ is interviewed by a teacher, $A$ fails to answer 70 questions.\\
On analysis, we see that $A$ has not truly learned the concept in $Q_1$; instead, $A$ solves questions by relying on common question patterns seen in $Q_1$, and associating them with the provided answers. To fix this, suppose $A$ is provided 1000 solved questions from $Q_2$. On testing, we find that $A$ now correctly answers 90/100 unsolved questions from $Q_2$, but only 55/100 from $Q_1$, and 35/100 in the interview. Now, we provide $A$ with 100 question banks in a similar manner, and find that $A$'s performance on both $Q_1$ and $Q_2$ is 70/100, and is 40/100 for the interview. To improve interview performance, suppose that the interviewer prepares an additional question bank $Q_i$, then if $A$ self-learns using both $Q_1$ and $Q_i$, then the scores for $Q_1$, $Q_2$, and the interview are 80, 45, and 80/100 respectively. However, if the interviewer changes, $A$ again fails to correctly answer 70 questions.\\
Since the provision of additional question banks in different settings was not very effective, we introduce a set of constraints in $A$'s self-learning strategy that disallows $A$ from picking up on questions patterns and answer associations. However, these constraints increase the time that $A$ spends on self-learning, and the number of question banks required (also, in turn the money spent if question banks are rented on a time basis). We find that this improves $A$'s accuracy in answering $Q_1, Q_2$, and interview questions to 95, 70, and 50/100 respectively. \\
Clearly, the above methods do not fully solve the problems in $A$'s self-learning strategy. This leads us to question where the problem actually lies-- \textit{is it in the learning strategy or in the learning material?} Intuitively, improving $A$'s learning strategy is conducive only if $A$ is being provided high quality learning material without any scope for identifying question patterns and answer associations; this forces $A$ to look beyond idiosyncrasies of the question banks that $A$ is tested on. \\
\textit{How do we create high quality datasets for models to learn from?} To define `quality', we require a \textit{quality index for machine learning}, similar to those used in the domains of power~\cite{bollen2000understanding}, air~\cite{jones1999indoor}, food~\cite{grunert2005food} and water~\cite{world1993guidelines}. Recently, based on a broad survey of AI literature, DQI \cite{Mishra2020DQIMD} has been proposed as a data quality index for NLP; here, relevant text properties that lead to either spurious or inductive bias are identified and used to construct a formula that quantitatively evaluates benchmarks.\\
DQI comprises of 133 terms; in this work we aim to study how some of these terms both individually and collectively can help models learn tasks in a few-shot setting. We aim to conduct two types of experiments: (i) dataset pruning on existing benchmarks using different DQI terms (individually and/or combined), and (ii) controlled crowdsourced creation of high quality datasets based on DQI. 
\vspace{-3mm}
\section{Related Work:}
\vspace{-3mm}
\label{sec2}

Our progress in AI is evaluated by building and solving increasingly harder benchmarks. This in turn leads to the development of new models and architectures. This trend requires heavy resource investment, in terms of time, cost, hardware, etc. However, in this process, we must ask if we can \textit{truly rely on our benchmarks}. A series of recent works have shown that models exploit spurious bias -- unintended correlations between input and output \cite{torralba2011unbiased, bras2020adversarial}-- to solve tasks, instead of actually learning the task from underlying data features \cite{gururangan2018annotation, schwartz2017effect, poliak2018hypothesis, tsuchiya2018performance, kaushik2018much, tan2019investigating,le2020adversarial, mishra2021robust}. \\
The mitigation of spurious bias has consequently become an increasingly prevalent track of research. Some of the most common methods to achieve this are dataset pruning \cite{sakaguchi2019winogrande,li2019repair,li2018resound,wang2018dataset}, residual learning \cite{clark2019don,he2019unlearn,mahabadi2019simple},  adversarial dataset creation \cite{zellers2018swag,nie2019adversarial}, and counterfactual data augmentation \cite{kaushik2019learning,gardner2020evaluating}. \\
Each of these methods focuses on a specific part of the data-model loop: (i) accepting/ rejecting a data sample created by a crowd-worker \cite{nie2019adversarial}, (ii) retaining/ removing data with adversarial filtering \cite{sakaguchi2019winogrande,li2019repair,li2018resound}, (iii) augmenting only counter factual data \cite{kaushik2019learning,gardner2020evaluating}, and/or (iv) including data only if it can fool the model \cite{zellers2018swag,nie2019adversarial}; they are all commonly limited by binary evaluation, and can also introduce new kinds of bias, overfitting to a specific model or task \cite{liu2019inoculation}.\\
Binary evaluation in particular, is extremely restrictive as it only allows inclusion or deletion of data, and further appends an overhead on human evaluators as there is uncertainty in class distinction. Some other limitations include: (i) resource wastage in the initial creation of `biased' data, (ii) dataset creators are likely to repeat mistakes that lead to the making of biased data, as they do not learn what constitutes biased data, (iii) important aspects of bias-- such as its dependency on a train-test split-- are ignored, (iv) model training on each iteration increases time complexity, and (v) there is too much effort required on the part of crowdworkers/authors/experts without providing a suitable and/or illustrative feedback channel to educate data creators. \\
Using DQI to quantify benchmark quality can potentially address these issues; higher DQI implies lower bias and higher generalization.
\vspace{-2mm}
\section{Method}
\vspace{-2mm}
We intend to utilize DQI in two ways: (i) dataset pruning, and (ii) dataset creation. We start by addressing if \textit{\textbf{(H1)} we can learn a task effectively with smarter sample selection(pruning).} However, pruning overlooks resource wastage in creating biased data. So, we investigate if \textit{\textbf{(H2)} we can leverage our pruning approach to assist crowd workers in constructing a smaller, but higher quality dataset in the first place, such that pruning is no longer required.} Answering \textbf{H2} will justify the utility of our question--\textit{Is high quality data all we need?}-- and change the deep learning trend of creating big datasets.
\vspace{-2mm}
\subsection{Dataset Pruning:}
\vspace{-2mm}
\textit{Do we really need big datasets? \cite{mishra2020we}} Motivated by the process of human learning which relies on deep background knowledge about the world-- we don't need access to hundreds of online materials to learn a topic, rather we intentionally avoid many noisy, distracting, and irrelevant materials-- we probe this question.  Considering that pre-training on large datasets has imparted linguistic knowledge to models like BERT \cite{devlin2018bert} and RoBERTA \cite{liu2019roberta}, we realize that models no longer need to learn from scratch; instead, learning task-specific terminology (such as `Entailment'/`Neutral'/`Contradiction'labels for Natural Language Inference) suffices, and might not necessitate the use of large datasets.\\
We therefore aim to find the high quality subset of benchmark data required to learn a task. Our approach is inspired by human tendency to: (i) estimate the presence of relevant materials from the total available material, (ii) remove redundant/irrelevant/known content from the initially selected material, and (iii) use background knowledge of the task, task priority, and time available for learning to heuristically sort and select relevant (i.e., high quality) content. \\
\textbf{Algorithm:} We mimic this material selection process in Algorithm \ref{algo:one}. We use 2 modules for learning-- (i) AFLite \cite{bras2020adversarial, sakaguchi2019winogrande}, and (ii) DQI. AFLite is a recent technique for adversarial filtering of dataset biases using linear models, whereas DQI has a method to quantify quality of samples, with or without annotation. \\
\textbf{Formalization:}  Let $M$ be the model, full dataset $D$ and pruned dataset $S$, and for each sample $s$, $E(s)$: evaluation score,$C(s)$: correct evaluation score, and $P(s)$: predictability score.

\begin{algorithm}
\small
\SetAlgoLined
\KwResult{Input: Dataset $D$ and Models $M$:[Logistic Regression, SVM]; Hyper-Parameters $b$, $m$, $n$, $t$ and $tau$; Output: Pruned dataset $S$}
a=0\;
\For {$a<100$}
{Select $a$\% random samples from $D$ and let $acc$ be IID accuracy of model $M$ at iteration $x$\;
\eIf {$acc(x)>acc(x-1)$}
{a=a+b}
{a=a}}
$D$= $a$\% of samples from $D$\;
Get $D$'s embeddings by finetuning RoBERTA on 10\% of $D$ and discard this 10\%\;
$S=D$\;
$E(s)=0$ and $C(s)=0$ for all $s$ in $S$ \;
\While {||S|| > n} {
 \ForAll{$i \in m$}{
 Randomly select trainset of size $t$ from $S$ and let $y$ =0\;
 \While {$y$ < 2}
 {
 Train $M$[$y$] on $t$ and evaluate on rest of $S$ i.e. $V$ \;
 \ForAll{$s \in V$}{ $E(s) = E(s)+1$\; \If {model prediction is correct}
 {$C(s) = C(s)+1$}}
 $y=y+1$
 }
 }
  \ForAll{$s \in S$}{$P(s)=C(s)/E(s)$}
  Shortlist instances where $P(s) > tau$ \;
  Sort shortlisted instances based on DQI values and delete $k$ lowest DQI instances
 }
 \caption{High Quality Sample Selection}
 \label{algo:one}
 \end{algorithm}
 
\vspace{-2mm}
\subsection{Dataset creation:}
\begin{figure}
\centering
    \includegraphics[width=\textwidth]{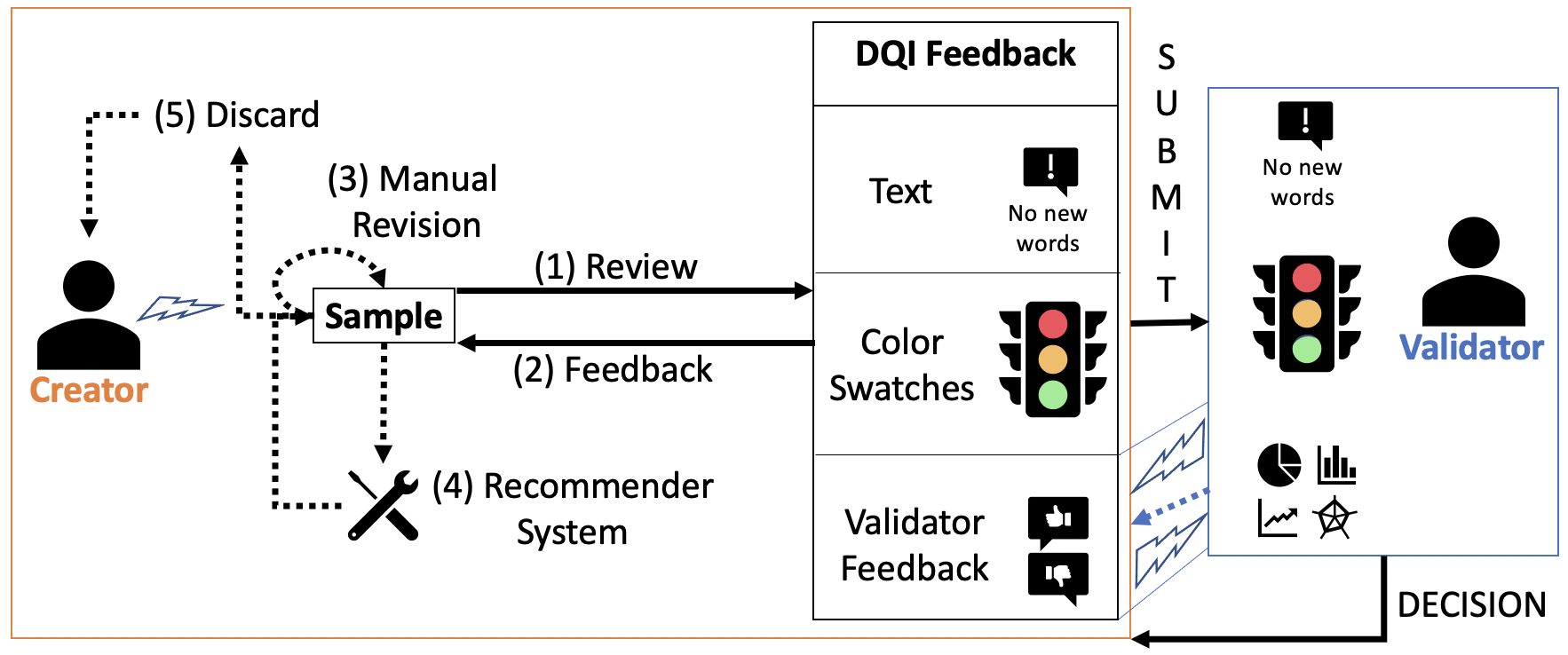}
    \caption{Data creation workflow. Creators have the choice of manually revising, fixing, or discarding samples. Validators may choose to provide feedback along with the decision to accept or reject samples. Dotted lines indicate steps where user choices are available.}
    \label{fig:vaidaworkflow}
\end{figure}

\vspace{-2mm}
\label{sec3.2}
While dataset pruning verifies the hypothesis that \textit{we don't need big datasets to learn a task when using pre-trained models}, other problems associated with pruning (Section \ref{sec2}), particularly resource and time wastage involved in creating biased data, remain unaddressed. We therefore plan to implement a DQI-in-the-loop approach as proposed in a recent work \cite{arunkumar2020real}, to recreate datasets and answer our second question-- \textit{Is a dataset created using DQI in a crowdsourcing setup equivalent to/better than one obtained via pruning?}\\
We propose a crowdsourcing workflow, as shown in Figure~\ref{fig:vaidaworkflow}. This workflow will support data creators; newly created samples are evaluated by DQI, and feedback is given to the user about potential biases. Feedback can be shown to indicate if a particular aspect of the data created might lead to spurious bias-- encouraging sample modification to decrease the presence of artifacts and increase generalization capability-- using: (i) component-wise text feedback on specific sample characteristics, (ii) color-swatches, with easy-to-interpret traffic signal color coding (red, yellow, and green), and (iii) feedback from a data validator. We also intend to provide a recommender system based on DQI to further assist creators in converting red signals to green. We will experiment with these modules over varying granularities-- i.e., feedback is provided at DQI component/sub-component/term levels. After potentially iterative cycles of feedback and revision, the sample will finally be submitted for bechmark inclusion.\\
Validators will evaluate submitted samples across different granularities in order to examine how each individual sample contributes to the overall quality of the current dataset state. This will utilize the first two feedback conditions presented to creators, along with visualizations, based on relevant text characteristics considered in each DQI component. The validators will then communicate sample decisions to the creators, with additional feedback (compulsory when the sample is rejected). This will enable continuous feedback about creator performance during creation, and also support data validators.\\
This framework aids in the creation of a new high quality dataset, and also enables opportunities for various novel applications (such as dataset refurbishment), inspiring the next generation of datasets and models.

\vspace{-2mm}
\section{Experimental Protocol}
\vspace{-2mm}
\label{sec 4}

We perform an initial exploratory analysis and have promising results for dataset pruning. This is in line with our three step approach to mimic human learning. We specifically address the third step-- imparting background knowledge and heuristics in learning-- and test 1 DQI term. 

\vspace{-2mm}
\subsection{Exploratory Analysis:}
\vspace{-2mm}
In our preliminary experiments \cite{mishra2020we} (Table \ref{tab1}), we utilize the first term of DQI $C_1$ (component 1) to prune SNLI \cite{bowman2015large} to $\sim 1-2\%$ of its original size (550K). When RoBERTA is trained with our pruned dataset, it achieves near-equal performance on the SNLI dev set, as well as competitive zero-shot generalization on: (i) NLI Diagnostics \cite{wang2018glue}, (ii) Stress Tests \cite{naik2018stress}, and (iii) Adversarial NLI \cite{nie2019adversarial}. This indicates that we might not need big datasets to learn a task. 

\begin{table}[!htb]
\scriptsize
    \begin{minipage}{.15\linewidth}
      \centering
        \begin{tabular}{@{}cc@{}}
\toprule
Size&IID\\(Random) &  \\ \midrule
5000  & 36.77 \\
10000 & 77.45 \\
15000 & 81.69 \\
        \end{tabular}
    \end{minipage}%
    \begin{minipage}{.9\linewidth}
      \centering
      \resizebox{0.9\linewidth}{!}{
        \begin{tabular}{@{}cccccccccccc@{}}
\toprule
Size  & IID   & \multicolumn{3}{c}{OOD ANLI} & \multicolumn{4}{c}{OOD NLI Diagnostics} & \multicolumn{3}{c}{OOD Stress Combined} \\ \midrule
      &       & R1      & R2      & R3       & Knowl.    & LS      & Logic   & PAS     & Comp.     & Distraction     & Noise     \\ \midrule
550k  & \textbf{89.64} & 36.6    & 30.5    & 31.33    & \textbf{57.64}     & \textbf{62.23}   & 53.8    & 66.51   & \textbf{51.63}     & 72.13           & \textbf{79.52}     \\\midrule
5000  & 87.47 & 32.6    & 31.8    & 28       & 50.35     & 61.14   & 48.37   & \textbf{67.45}   & 35.29     & 65.72           & 73.97     \\
10000 & 87.93 & 34.5    & \textbf{33}      & \textbf{31.67}    & 55.9      & 61.14   & 53.26   & 66.75   & 45.94     & \textbf{74.88}           & 74.62     \\
15000 & 88.95 & \textbf{37.2}    & 28.3    & 29.17    & 56.6      & 56.79   & \textbf{54.62}   & 65.8    & 45.94     & 70.66           & 77.71            \end{tabular}}%
    \end{minipage} 
    \caption{Left-- Random Selection. Right-- Pruned set results. Highlighted points: best performances.}
    \label{tab1}
\end{table}
\vspace{-2mm}
\subsection{Proposed Experiments:}
\vspace{-2mm}
We plan to further investigate high quality selection criterion by performing full scale pruning experiments on the SNLI\cite{bowman2015large} and MNLI~\cite{williams2017broad}, and SQUAD 1.1~\cite{rajpurkar2016squad} datasets. MNLI and SNLI will use the same OOD datasets as in Table \ref{tab1}. For SQUAD, we will use NewsQA\cite{trischler2016newsqa}, TriviaQA\cite{joshi2017triviaqa}, SearchQA\cite{dunn2017searchqa}, HotpotQA\cite{yang2018hotpotqa}, and Natural Questions\cite{kwiatkowski2019natural} as OOD datasets, in line with a recent work \cite{kamath2020selective}. We will also be recreating SNLI, MNLI, and SQUAD 1.1 using DQI-in-the-loop as demonstrated in in our workflow (Figure \ref{fig:vaidaworkflow}).\\
\textbf{Pruning Experiments:} We intend to prune based on additional terms in DQI \cite{Mishra2020DQIMD}. DQI is calculated based on 7 components, 20 sub-components, and 133 terms. In order to short-list sub-components that we can reasonably expect to be useful, we will first conduct pruning on SNLI based on all 7 components, individually. We will compare IID accuracy for various pruned sizes (Table \ref{tab1}), and shortlist the two components that result in the highest IID accuracy of the pruned set. We will then prune with the terms of selected components (based on initial SNLI pruning), to varying sizes (similar to Table \ref{tab1}), to find out which DQI terms result in achieving higher IID accuracy, over all 3 datasets.\\
We also plan to  perform an ablation study of the DQI, AFLite and Coarse Action (Algorithm \ref{algo:one} lines 2-9) modules, by removing them from the algorithm, on the shortlisted components. In all these experiments our pruning happens purely based on IID test set accuracy. Zeroshot OOD evaluation is just done to ensure that the pruned dataset does not contribute mainly spurious bias.\\
In RoBERTA, we plan to change the learning rate from 1e-6 to 1e-5, and vary $b$ as 100, 1000, 2000, and 5000. $n$ is the target dataset size, and $t$ is the training set size; we plan to vary both from 10 \% of $S$ (the pruned dataset) to 75\% of $S$, in 15\% increments. $m$ will be varied from 8 to 124 in increments of either 16 or 32. Other hyperparameters will be fixed as per Hugging face transformers\cite{wolf2019transformers}.\\
\textbf{Expectations:} In DQI, terms are synonymous with sub-components, except for $C_2$ and $C_6$, as these two components address quality at word, POS tag (adjective, adverb, noun, and verb), bigram, trigram, and sentence granularities; $C_6$ further calculates terms label-wise, which we will ignore for the purpose of pruning (-80 terms). If $C_2$ and/or $C_6$ are shortlisted based on component-wise pruning, they will contribute 8 and 40 terms respectively. In other cases, components will contribute 1-5 terms. We will prune all 3 datasets with the terms of selected components (based on initial SNLI pruning), to varying sizes, similar to Table \ref{tab1}. In recent work, word overlap \cite{gururangan2018annotation,bras2020adversarial} and semantic textual similarity \cite{mishra2020our} have been dominant in producing spurious bias; we therefore expect to shortlist $C_3$ and $C_5$ in our component-wise experiments.\\
Previous work has found that the amount of artifacts in datasets is in the order: SNLI>SQUAD>MNLI \cite{gururangan2018annotation,bras2020adversarial,tsuchiya2018performance,tan2019investigating,poliak2018hypothesis,Mishra2020DQIMD}. Accordingly, we expect the size of the equivalent pruned set (2\% for SNLI) to be in reverse. Additionally, considering the human motivation for Algorithm \ref{algo:one}, we expect our ablation experiments to affect performance in the order of DQI>AFLite>Coarse Action, with reverse order for effect on pruning time.\\
\textbf{Creation Experiments:} We will use our creation workflow (Section \ref{sec3.2}) in crowdsourcing, to recreate SNLI, MNLI, and SQUAD 1.1. For each respective dataset (without pruning), we will select the smallest pruned dataset size that results in IID accuracy within +/-5\% of the original IID accuracy and create a similar size data using crowdsourcing setup. We will additionally perform ablation studies with subsets of creators, across the different quality feedback methods.  We will be using the default hyperparameters mentioned in the DQI work \cite{Mishra2020DQIMD}.\\
\textbf{Expectations:} In the ablation studies, we expect number of samples, sample quality and IID/OOD performance to be affected in the following order: all feedback modes>recommender system>text feedback~validator feedback>color-swatches. We expect the time involved to follow the reverse order.

\bibliography{nips.bib}
\bibliographystyle{abbrvnat}
\end{document}